\documentclass{INTERSPEECH2023}

\interspeechcameraready 

\title{Inter-connection: Effective Connection between Pre-trained Encoder and Decoder for Speech Translation}
\name{Yuta Nishikawa$^1$, Satoshi Nakamura$^1$}
\address{$^1$Nara Institute of Science and Technology, Japan}
\email{nishikawa.yuta.oz3@is.naist.jp}

\begin{document}

\maketitle

\begin{abstract}
In end-to-end speech translation, speech and text pre-trained models improve translation quality. Recently proposed models simply connect the pre-trained models of speech and text as encoder and decoder. Therefore, only the information from the final layer of encoders is input to the decoder. Since it is clear that the speech pre-trained model outputs different information from each layer, the simple connection method cannot fully utilize the information that the speech pre-trained model has. In this study, we propose an inter-connection mechanism that aggregates the information from each layer of the speech pre-trained model by weighted sums and inputs into the decoder. This mechanism increased BLEU by approximately 2 points in en-de, en-ja, and en-zh by increasing parameters by 2K when the speech pre-trained model was frozen. Furthermore, we investigated the contribution of each layer for each language by visualizing layer weights and found that the contributions were different.
\end{abstract}

\section{Introduction}

Speech translation, also known as spoken language translation, is the process of converting spoken language from one language to another. The primary objective of speech translation is to enable seamless communication between people who speak different languages and it has numerous practical applications in fields such as international business and tourism.

In recent years, there has been significant progress in speech translation technology driven by advances in deep learning and spoken language processing. State-of-the-art speech translation systems use end-to-end modeling with large scale self-supervised learning (SSL) models of speech and text modalities. By combining SSL models of these different modalities, it is possible to efficiently build a speech translation model with a small amount of data. However, simply connecting SSL models alone does not fully extract the information processed by speech SSL models. It is known that semantic, phonological, and word-like information is embedded in the intermediate layer of the SSL speech model \cite{Pasad2021LayerWiseAO}.

In this paper, we propose the Inter-connection Mechanism, a method to fully extract information from the SSL model of speech and utilize it for speech translation. In addition, we also perform a layer-wise analysis of the SSL model and share our findings.

\section{Related Work}
In recent years, there has been a growing interest in using pre-trained models to improve the performance of speech translation systems. Pre-trained models are neural network models that have been trained on large datasets of unlabeled text or speech data and can be fine-tuned on a smaller task-specific dataset to achieve high levels of accuracy.

SSL models in speech \cite{Baevski2020wav2vec2A, Hsu2021HuBERTSS, Chen2021WavLMLS} and text modalities \cite{lewis-etal-2020-bart, liu-etal-2020-multilingual-denoising} have been shown to be useful for various downstream tasks, contributing significantly to performance improvements in speech recognition, speech translation, machine translation, and more. In particular, significant results have been achieved in speech translation by combining the speech SSL model with the text SSL model \cite{pham-etal-2022-effective, zhang-etal-2022-niutranss, tsiamas-etal-2022-pretrained}. However, the research to date has simply connected the two models and has not been able to fully extract the information from the speech SSL model.

In this study, we focus on how to connect the SSL models of speech and text, and propose a method to maximize the information possessed by the speech SSL model. In addition, we show that each language provides a different contribution to each layer.

\section{Model Architecture}
\begin{figure*}[th]
  \centering
  \includegraphics[width=400pt]{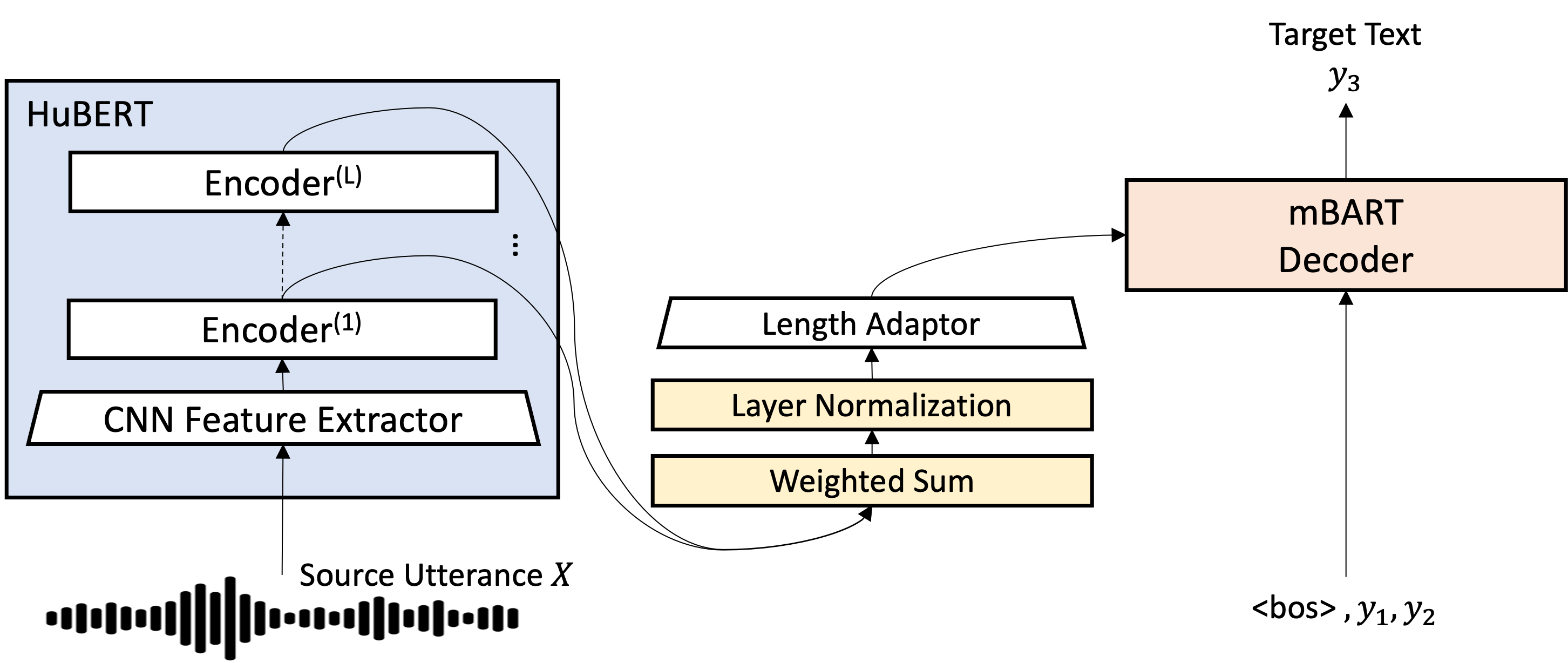}
  \caption{Inter-connection between pre-trained encoder and decoder}
  \label{fig:inter-connection}
\end{figure*}

We constructed our model following the model proposed by the UPC Translation Group at the International Conference on Spoken Language Translation (IWSLT) 2022 \cite{tsiamas-etal-2022-pretrained} as the baseline model.
In the baseline model, speech and text pre-trained models are connected together via the length adaptor.
We attempted to improve upon this baseline architecture by adding the Inter-connection Mechanism as shown in Figure \ref{fig:inter-connection}.

\subsection{Pre-trained Models}
Combining a pre-trained model of the speech modality with a pre-trained model of the text modality is key to achieving high translation quality for speech translation. Therefore, we use HuBERT (Hidden-unit Bidirectional Encoder Representations from Transformers) \cite{Hsu2021HuBERTSS} as the encoder and mBART50 \cite{liu-etal-2020-multilingual-denoising} as the decoder.

\begin{quote}
    \begin{itemize}
        \item \textbf{HuBERT} consists of a convolutional neural network (CNN) downsampler and multiple transformer encoder blocks. HuBERT Base has 12 layers of transformer blocks and HuBERT Large has 24 layers. The pre-trained model uses 60-k hours of unlabeled data as training data and is trained to output generic speech representations through self-supervised learning.
        \item \textbf{mBART50} is an encoder-decoder transformer model that is trained by denoising noisy text. It is also a multilingual model trained with 50 languages. In this study, only the decoder part is used.
    \end{itemize}
\end{quote}

These models are connected as a transformer encoder-decoder model to form one end-to-end model.

\subsection{Length Adapter}
Because we connect components of two different modalities together, the length of the output sequence from the encoder and the length of the sequence handled by the decoder are different. To bridge this gap, we downsample the output of the encoder using a length adaptor (LA). The length adaptor consists of a series of convolutional layers, which are better connected by matching the lengths of the series.

\subsection{Inter-connection}
The intermediate layer of HuBERT contains a wealth of information that is important for downstream tasks, such as phoneme and word recognition.
However, the simple connection method described above cannot fully exploit HuBERT's performance.
To efficiently utilize the information in the intermediate layer and input it to the decoder, we propose Inter-connection (Figure \ref{fig:inter-connection}).
This mechanism assigns weights to each transformer layer in the encoder and aggregates the tensors from the outputs of each layer.
The weights are additional learnable parameters optimized through the training.
The detailed formula is given in equation (\ref{equ:inter-connection}).

\begin{equation}
    \label{equ:inter-connection}
    \mathbf{\hat{H}}=LayerNorm(\sum^{L}_{l} \mathbf{H_\textit{l}} w_l)
\end{equation}

In this formula, $\mathbf{H_\textit{l}}$ denotes the output tensor of the \textit{l}-th encoder layer, $w_l$ represents the weights assigned to the \textit{l}-th encoder layer, and $\bold{\hat{H}}$ represents the output of the inter-connection.
We also apply layer normalization after the weighted aggregation to stabilize the training.

\section{Experiments}
\subsection{Experimental Settings}
\subsubsection{Datasets}
In this study, we trained a multilingual model that can translate three language pairs: English-German (en-de), English-Japanese (en-ja), and English-Chinese (en-zh). For training, we used six corpora, shown in Table \ref{tab:dataset}. MuST-C v2 and CoVoST-2 are used for en-de, en-ja, and en-zh. In addition, MuST-C v1, Europarl-ST, and TED-LIUM are used for en-de.
Basically, corpora in Group (a) are used for training the model.
Corpora in Group (b) are the same as the corpora used for training the baseline model \cite{tsiamas-etal-2022-pretrained} and are used for fair comparison.

\begin{table}[th]
  \caption{Training data measured in hours}
  \label{tab:dataset}
  \centering
  \begin{tabular}{llrrr}
    \toprule
    \textbf{Group} & \textbf{Dataset} & \textbf{en-de} & \textbf{en-ja} & \textbf{en-zh} \\
    \midrule
    (a)      & MuST-C v1 \cite{di-gangi-etal-2019-must}  & 408h  &       &      \\
    (a), (b) & MuST-C v2 \cite{di-gangi-etal-2019-must}  & 436h  & 526h  & 545h \\ 
    (a), (b) & Europarl-ST \cite{Europarl2020}           & 83h   &       &      \\
    (a), (b) & CoVoST-2 \cite{wang-etal-2020-covost}     & 413h  & 413h  & 413h \\
    (a)      & TED-LIUM \cite{Rousseau2012TEDLIUMAA}     & 415h  &       &      \\
    \hline \\ [-1.5ex]
             & Total                                     & 1755h & 939h  & 958h \\
    \bottomrule
  \end{tabular}
\end{table}

\subsubsection{Implementation Details}
We employed the transformer encoder-decoder architecture, where the encoder is initialized by HuBERT Large. The encoder consists of a 7-layer convolutional feature extractor and 24-layer transformer encoder. The model is trained on Libri-Light, a 60-k hour unlabeled speech dataset. The feature extractor has 512 channels with kernel sizes of 10, 3, 3, 3, 3, 2, and 2 and strides of 5, 2, 2, 2, 2, 2, and 2. 
The decoder is initialized with mBART50 and consists of a 12-layer transformer decoder.
Each layer in the transformer encoder and decoder has a dimensionality of 1024, feed-forward network dimension of 4096, 16 heads, rectified linear unit (ReLU) activations, and uses pre-layer normalization. The length adaptor between the encoder and decoder is a 3-layer convolutional network with 1024 channels, stride of 2, and uses gated linear unit (GLU) activations. The embedding layer and linear projection weights of the decoder are shared and have a size of 250,000.

The inputs to the model are waveforms with a 16-kHz sampling rate that are normalized to zero mean and unit variance.
During training, each source audio is augmented \cite{wavaugment} (before normalization) with a probability of 0.8. We train multilingual models on all the data of Table \ref{tab:dataset} with a maximum source length of 400,000 and target length of 1024 tokens. We use gradient accumulation and data parallelism to achieve a batch size of approximately 32 million tokens.
We use Adam \cite{Kingma2014AdamAM} with $\beta_1 = 0.99, \beta_2 = 0.98$ and a base learning rate of $2.5 \cdot 10^{-4}$.
The learning rate is controlled by a tri-stage scheduler with phases of 0.15, 0.15, and 0.70 for warm-up, hold, and decay accordingly, while the initial and final learning rate has a scale of 0.01 compared to base.
We used sentence averaging and gradient clipping of 20.
We applied a dropout of 0.1 before every non-frozen layer and use masking for 10-length time spans with a probability of 0.2, and masking for 20-length channel spans with a probability of 0.1 in the encoder feature extractor's output.
The loss is the cross-entropy loss with label smoothing of 0.2.

\subsubsection{Parameter Freezing Strategy}
We experimented under three types of freezing strategies: HuBERT freezing, LayerNorm and Attention (LNA) fine-tuning, \cite{tsiamas-etal-2022-pretrained} and full fine-tuning.
The baseline model employs LNA fine-tuning to save memory; we also built a model with parameters frozen in the LNA strategy for fair comparison with the baseline.

\subsection{Results}

\subsubsection{Translation Quality}
\begin{table*}[th]
    \caption{BLEU scores and BERTScores for MuST-C tst-COMMON set}
    \label{tab:translation-quality}
    \centering
    \begin{tabular}{lllccccccc}
        \toprule
        \multicolumn{4}{c}{} & \multicolumn{3}{c}{\textbf{BLEU}} & \multicolumn{3}{c}{\textbf{BERTScore}} \\
        & \textbf{Model} & \textbf{Frozen} & \textbf{Corpora} & 
        \textbf{en-de} & \textbf{en-ja} & \textbf{en-zh} &
        \textbf{en-de} & \textbf{en-ja} & \textbf{en-zh} \\
        \midrule
        (1-A) & HuBERT-mBART+LA                  & HuBERT & (a) & 24.68 & 11.86 & 20.55 & 0.577 & 0.763 & 0.467 \\
        (1-B) & HuBERT-mBART+LA+Inter-connection & HuBERT & (a) & \textbf{26.79} & \textbf{14.15} & \textbf{22.20} & \textbf{0.591} & \textbf{0.771} & \textbf{0.484} \\ 
        \hline \\ [-1.5ex]
        (2-A) & HuBERT-mBART+LA \cite{tsiamas-etal-2022-pretrained} & LNA & (b) & \textbf{29.27} & 14.89 & \textbf{24.84} & -     & -     & - \\
        (2-B) & HuBERT-mBART+LA+Inter-connection                    & LNA & (b) & 28.73 & \textbf{15.17} & 24.41 & 0.634 & 0.785 & 0.532 \\
        \hline \\ [-1.5ex]
        (3-A) & HuBERT-mBART+LA                  & None   & (a) & 30.48 & 15.81 & \textbf{24.82} & 0.644 & 0.786 & \textbf{0.541} \\
        (3-B) & HuBERT-mBART+LA+Inter-connection & None   & (a) & \textbf{30.67} & \textbf{16.22} & 24.59 & \textbf{0.647} & \textbf{0.786} & 0.539 \\
        \bottomrule
    \end{tabular}
\end{table*}

We experimented with evaluating translation quality by building multilingual models with the three language pairs, en-de, en-ja, and en-zh using bilingual evaluation understudy (BLEU) \cite{papineni-etal-2002-bleu} and BERTScore \cite{bert-score}.
Our main results are shown in Table \ref{tab:translation-quality}.
We used a model in which HuBERT and mBART were connected by a length adaptor as a baseline \cite{tsiamas-etal-2022-pretrained} and investigated whether adding inter-connection improves translation quality.
We also compared HuBERT with all frozen parameters and all fine-tuned parameters.
Additionally, for fair comparison, we compared the LNA fine-tuned model trained by corpora belonging to Group (b) in Table \ref{tab:dataset}.

The results showed that BLEU increased 2.11 in en-de, 2.31 in en-ja, and 1.65 in en-zh, and BERTScore increased slightly overall when the parameters of HuBERT were frozen.
The overall increase shows that a slight increase in the parameter numbers significantly improves the performance of the model.
When parameters were frozen by LNA strategy, BLEU increased 0.28 in en-ja and decreased 0.54 in en-de and 0.43 in en-zh.
When all parameters were fine-tuned, BLEU increased 0.19 in en-de and 0.39 in en-ja and decreased 0.23 for en-zh. The BERTscore increased slightly in en-de and en-ja and decreased slightly in en-zh.

In en-zh, the increase in performance when all parameters were frozen was smaller than those in the other language pairs, and performance decreased when all parameters were fine-tuned.
This seems to be due to interference between the importance of the layers required for en-zh and the importance of the layers required for the other language pairs.

In addition, we attempted to build a bilingual model for each language; however, the small amount of data resulted in overfitting and significantly degraded performance.

\subsubsection{Parameter Size Analysis}

The next question is how well it works for the parameter size.
We performed an experiment under three conditions on the model in which all HuBERT parameters were frozen.
Models of (1-A) and (1-B) from the conditions listed in Table \ref{tab:translation-quality} and the model with one new transformer block added to the final layer of HuBERT were compared with BLEU and the number of parameters.
Adding one more transformer block increased the BLEU by 0.71 points in en-de.
This result is smaller than the increase with the addition of 2.11 points of inter-connections.
Therefore, adding inter-connections is more efficient than simply increasing the number of parameters in the model.
Furthermore, in terms of the parameter size, adding one transformer block increased the parameter size by 12 M while adding an inter-connection increased the parameter size by only 2 K.
This result indicates that inter-connection is efficient in two aspects: performance and parameter size when freezing the pre-trained model.

\begin{figure}[th]
    \centering
    \includegraphics[width=\linewidth]{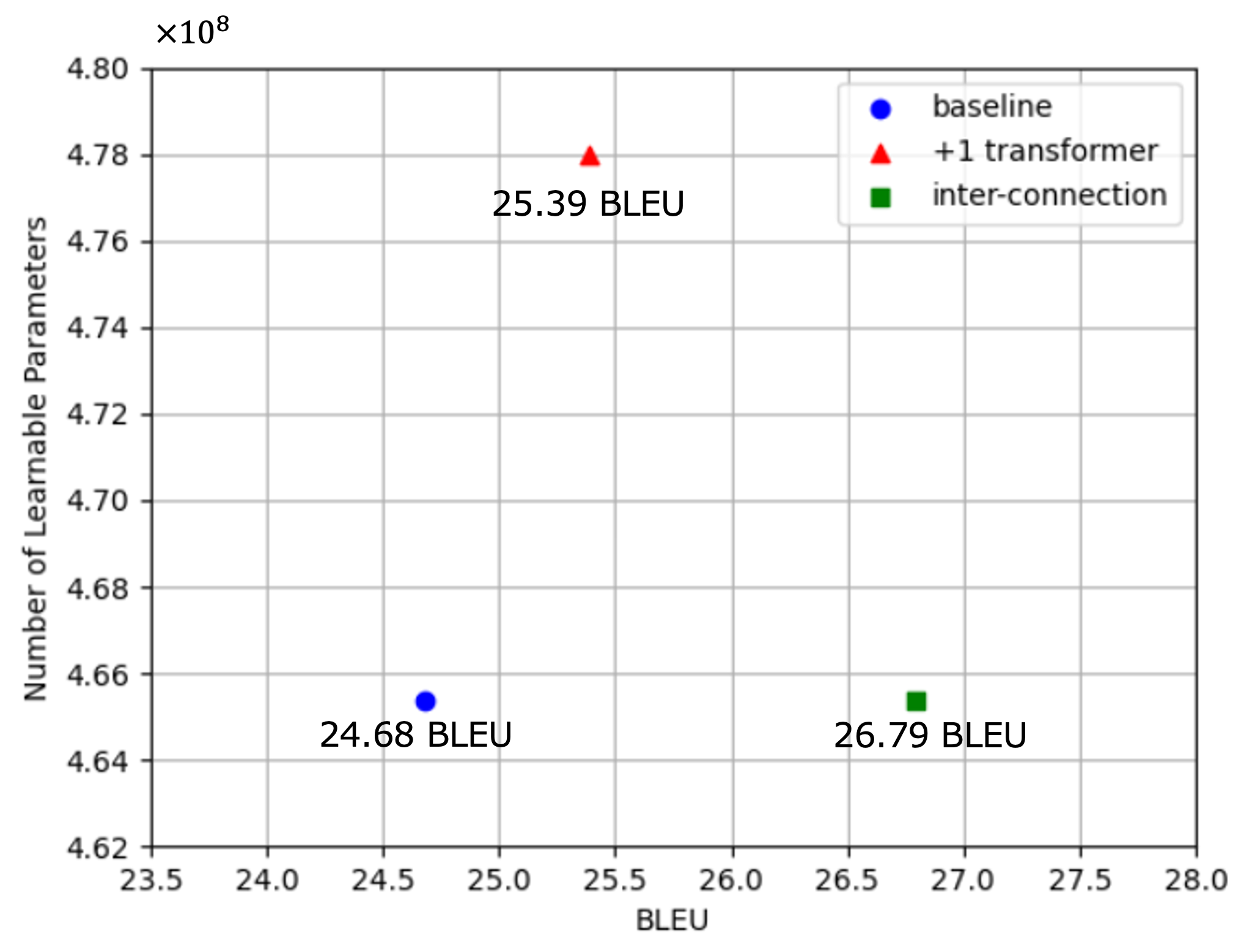}
    \caption{Model-by-model comparison with parameter size and BLEU scores in en-de}
    \label{fig:parameter-size-and-bleu}
\end{figure}

\subsubsection{Layer-wise Analysis}

\begin{figure}[h]
  \centering
  \includegraphics[width=\linewidth]{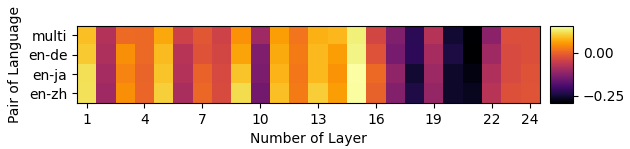}
  \caption{Weights of lang-specific models}
  \label{fig:lang-specific-weights}
\end{figure}
\begin{figure}[h]
    \centering
    \includegraphics[width=\linewidth]{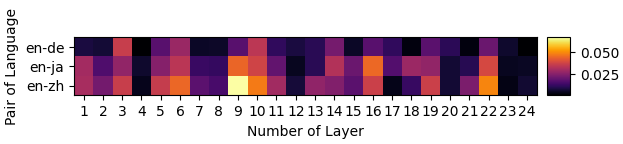}
    \caption{Difference from the multilingual model of weights on lang-specific models}
    \label{fig:lang-specific-weights-diff}
\end{figure}

We investigated whether the required information is different for each language. For this purpose, we trained lang-specific bilingual models for each language pair en-de, en-ja, and en-zh under the condition of HuBERT freezing.
Bilingual models were trained using the corpus shown in the Table \ref{tab:dataset} for each language.
The training conditions were the same as those described in section 4.1.2.
Additionally, we visualized the weight of each layer of inter-connection in Figure \ref{fig:lang-specific-weights} and calculated differences in the layer weights from the multilingual model (1-A) in Table \ref{tab:translation-quality} in Figure \ref{fig:lang-specific-weights-diff}.
The difference between the multilingual model and the bilingual model was computed using the equation (\ref{equ:weight-difference}).

\begin{equation}
    \label{equ:weight-difference}
    W_{diff} = |W_{multi} - W_{bi}|
\end{equation}

$W_{multi}$ represents the weight assigned to each encoder layer, and $W_{bi}$ represents the weight assigned to each encoder layer. 
By calculating the absolute value of the difference between these weights $W_{diff}$, we obtain the difference in weights.

We also calculated the cosine similarity of layer weights with the multilingual model to quantify the similarity with the multilingual model in Table \ref{tab:cossim-with-multilingual}.

\begin{table}[h]
  \caption{Cosine similarities of lang-specific models with a multilingual model}
  \label{tab:cossim-with-multilingual}
  \centering
  \begin{tabular}{lr}
    \toprule
    \textbf{Model} & \textbf{Cosine Similarity} \\
    \midrule
    en-de & 0.9906 \\
    en-ja & 0.9792 \\ 
    en-zh & 0.9694 \\
    \bottomrule
  \end{tabular}
\end{table}

As a result, we obtained weights that are close to the multilingual model. However, the cosine similarity with the multilingual model in en-ja and en-zh is lower than in en-de. This is partly due to the larger amount of training in the en-de data than in en-ja and en-zh. On the other hand, this result also indicates that the importance of the necessary information differs between languages. Therefore, training a multilingual model using inter-connection is hampered by the different importance of each layer between languages. In particular, in en-zh, where the most significant weight differences are observed in Figure \ref{fig:lang-specific-weights-diff}, the increase in performance is small under HuBERT freezing conditions and performance is decreased under full fine-tuning.
It is assumed that this phenomenon is caused by differences in language pronunciation and grammar.

\section{Conclusions}
In this study, we constructed an end-to-end speech translation model with pre-trained models connected with inter-connection.
The inter-connection is a weighted sum aggregation of hidden states from intermediate layers of the encoder to maximize the use of the pre-trained encoder's information.
As a result, the translation accuracy of the three language pairs, en-de, en-ja, and en-zh, increased by approximately 2 BLEU when the parameters of the pre-trained encoder were frozen.
We also found that performance can be increased more efficiently than simply stacking layers and increasing the parameter size.
In our future work, we will consider ways to prevent the overfitting that occurs when full fine-tuning on small data sets. We will also investigate the effectiveness of this method in other downstream tasks.

\section{Acknowledgements}
This work is supported by JSPS KAKENHI Grant Number JP21H05054.

\bibliographystyle{IEEEtran}
\bibliography{mybib}

\end{document}